\documentclass[10pt,twocolumn,letterpaper]{article}

\usepackage[pagenumbers]{cvpr} 

\usepackage{multirow}
\usepackage[table]{xcolor}

\definecolor{cvprblue}{rgb}{0.21,0.49,0.74}
\usepackage[pagebackref,breaklinks,colorlinks,allcolors=cvprblue]{hyperref}

\usepackage{booktabs}
\usepackage[table]{xcolor}
\usepackage{comment}

\definecolor{perframe}{RGB}{235,235,235}   
\definecolor{selfsup}{RGB}{222,235,247}    
\definecolor{videogen}{RGB}{224,242,219}   
\definecolor{expert3d}{RGB}{238,228,246}   

\definecolor{perframeText}{RGB}{90,90,90}      
\definecolor{selfsupText}{RGB}{35,76,120}      
\definecolor{videogenText}{RGB}{40,110,40}     
\definecolor{expertText}{RGB}{96,60,130}     

\newcommand{\perframetxt}[1]{\textcolor{perframeText}{#1}}
\newcommand{\selfsuptxt}[1]{\textcolor{selfsupText}{#1}}
\newcommand{\videogentxt}[1]{\textcolor{videogenText}{#1}}
\newcommand{\experttxt}[1]{\textcolor{expertText}{#1}}

\newcommand{\best}[1]{\textbf{#1}}
\newcommand{\secondbest}[1]{\underline{#1}}

\title{How Much 3D Do Video Foundation Models Encode?}

\author{Zixuan Huang$^{1*}$ \quad Xiang Li$^{1*}$ \quad Zhaoyang Lv$^{2}$ \quad James M. Rehg$^{1}$\\
$^1$University of Illinois at Urbana-Champaign, $^2$Impossible, Inc.
}

\begin{document}

\twocolumn[{%
\renewcommand\twocolumn[1][]{#1}%
\maketitle
\begin{center}
\vspace{-7mm}
\textbf{\url{https://vidfm-3d-probe.github.io/}}
\end{center}
\begin{center}
    \centering
    \captionsetup{type=figure}
    \includegraphics[width=1.0\textwidth]{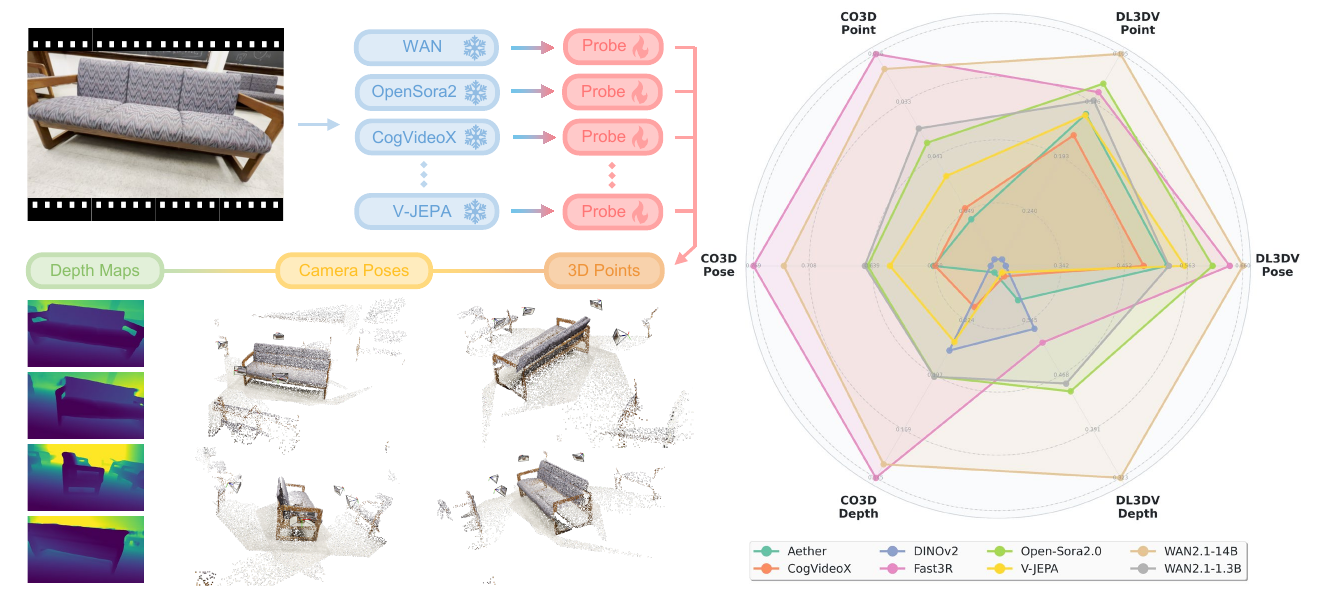}
    \captionof{figure}{We study the emergence of 3D in video foundation models by probing their features with 3D reconstruction tasks. Our study reveals state-of-the-art video generators develop strong 3D understanding even compared to 3D experts, despite only trained on 2D data.}
\label{fig:teaser}
\end{center}%
}]
\def\thefootnote{*}\footnotetext{Both authors contributed equally to this work.}\def\thefootnote{\arabic{footnote}}
\begin{abstract}
Videos are continuous 2D projections of 3D worlds. After training on large video data, will global 3D understanding naturally emerge? We study this by quantifying the 3D understanding of existing Video Foundation Models (VidFMs) pretrained on vast video data. We propose the first model-agnostic framework that measures the 3D awareness of various VidFMs by estimating multiple 3D properties from their features via shallow read-outs. Our study presents meaningful findings regarding the 3D awareness of VidFMs on multiple axes. In particular, we show that state-of-the-art video generation models exhibit a strong understanding of 3D objects and scenes, despite not being trained on any 3D data. Such understanding can even surpass that of large expert models specifically trained for 3D tasks. Our findings, together with the 3D benchmarking of major VidFMs, provide valuable observations for building scalable 3D models.
\end{abstract}    
\section{Introduction}

Recovering 3D structure from 2D visual observations is a long-standing research problem in computer vision, with
broad applications in AR/VR and embodied AI. Despite significant progress, the availability of high-quality 3D data at scale remains the bottleneck for current data-driven approaches. This fundamentally limits the scaling of 3D foundation models and makes it questionable whether we can learn truly generalizable models primarily from 3D data.

Compared to native 3D assets, videos are much easier to acquire at scale, with multiple large curated datasets already available~\cite{bain2021frozen,abu2016youtube,miech2019howto100m,chen2024panda70m}. The diversity and complexity of video data, with the fact that videos are 2D projections of 3D worlds, lead to a promising pathway for scalable 3D learning. Recent works study how to utilize video models for 3D, either by adding 3D control~\cite{he2024cameractrl,he2025cameractrl,bahmani2025ac3d,xu2024camco,sun2024dimensionx} or by producing 3D caches/estimations~\cite{team2025aether,lu2025matrix3d,zhang2025world,huang2025jog3r,mai2025can,jiang2025geo4d,yu2024viewcrafter,gu2025diffusion,ren2025gen3c,huang2025voyager,wu2025video,liang2024wonderland,zhang2025spatialcrafter} along with the original frame synthesis target. These works suggest that video priors are useful for 3D, but 3D-inconsistency artifacts, the requirement of 3D fine-tuning, and task-specific engineering leave it unclear whether video data alone can induce strong 3D awareness in a general-purpose setting. These confounds motivate a direct, model-agnostic evaluation.

In this paper, we present the first model-agnostic framework to probe the 3D awareness of video foundation models (VidFMs) pretrained on large-scale video data. We ask whether VidFMs develop internal representations of 3D structure and ego-motion and, if so, how strong and practically useful these representations are. We operationalize this question along four axes: 1) \textbf{Extent}: how does the 3D awareness of VidFMs compare to that of image models or specialized 3D models? 2) \textbf{Factor}: Which factors impact 3D awareness? Here, we focus on the effects of temporal reasoning, 3D finetuning and model scaling. 3) \textbf{Localization}: In which network layers, and at which timesteps in diffusion models, is this 3D information most concentrated? 4) \textbf{Implication}: Under limited resources (3D data and compute), can VidFM features be practically useful for 3D reconstruction tasks?

We posit that if a video model understands 3D worlds, it should be feasible to extract accurate 3D properties using shallow readout modules in a feedforward manner, without any post-optimization or fine-tuning of the base model. Unlike prior works that evaluate image models using depth and cross-view consistency~\cite{el2024probing}, or per-scene optimization with off-the-shelf initialization~\cite{chen2025feat2gs}, our proposed shallow feedforward readouts that estimate different 3D attributes from VidFMs' feature space are a more direct probe of globally consistent 3D properties from pretrained video models.

Specifically, we extract frozen spatialtemporal features from VidFMs, and design a probe model that predicts 3D points, camera poses and depth maps from these features. The probe model is a shallow VGGT~\cite{wang2025vggt}-like transformer, consisting of four alternating-attention layers and three read-out heads: two dense prediction heads for 3D points and depth maps, and one camera head. We train the probe model on top of various video features, including features extracted from self-supervised video models and video generation models of different performance and sizes. We measure the performance of point, camera and depth reconstruction as indicators of 3D awareness. 

Our study leads to the following novel findings:
\begin{itemize}
    \item \textbf{Extent}: Frontier video generation models exhibit great understanding of 3D objects and scenes, and can be close to or better than models trained with 3D data;
    \item \textbf{Factor \#1:} Temporal reasoning is critical to the formation of global 3D understanding;
    \item \textbf{Factor \#2:} Finetuning video generation models with 3D objectives improves 3D awareness on in-domain data, but may hurt generalization beyond data domains;
    \item \textbf{Factor \#3:} Model scaling leads to mixed impact on 3D awareness, with WAN2.1-14B performing significantly better than WAN2.1-1.3B, while CogVideoX-5B is slightly worse than CogVideoX-2B.
    \item \textbf{Localization:} The best layer and timestep to extract 3D-aware features are surprisingly consistent across all tested video diffusion models: mid-layer features with early-but-not-first timesteps lead to the highest 3D awareness.
    \item \textbf{Implication:} We implement and train a VGGT model using frozen VidFM (WAN2.1-14B) features. On CO3Dv2 and DL3DV, it consistently outperforms the standard DINO-based VGGT, indicating VidFM features may be better suited for 3D reconstruction under limited 3D data.
\end{itemize}

In summary, we conduct the first systematic model-agnostic evaluation on the 3D awareness of VidFMs and conclude with meaningful findings across multiple axes that the prior work has not surfaced well. Our findings are based on a comprehensive benchmark that compares 3D awareness of various video models, which can benefit the development of VidFMs by enabling the evaluation of their emerging 3D properties in a general-purpose way.
\section{Related Works}

\begin{figure*}[ht]
\centering
\includegraphics[width=0.95\textwidth]{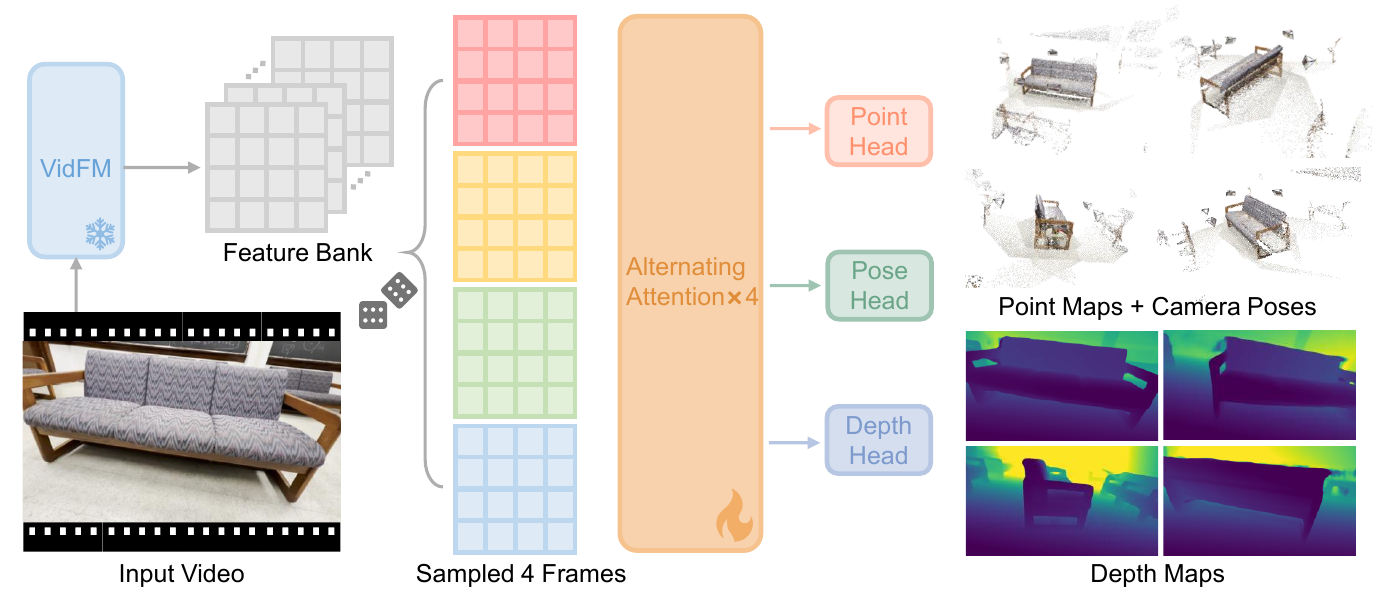}
\caption{\textbf{Overview of the Probe.} We extract video features using various video foundation models and keep the features frozen. We sample four frames from the original video clip and fetch the corresponding feature maps from the video features. We train the probe by taking these spatial features as input, and task the probe to estimate point maps, depth maps and camera poses. Our probe model consists of a shallow transformer and three readout heads. We measure the estimation errors as the main indicators of 3D awareness.}
\label{fig:ch6-arch}
\end{figure*}

\paragraph{Video foundation models (VidFMs)} are deep models trained on massive video data that achieve strong performance across various downstream tasks. Early works in the field employ self-supervised learning~\cite{bardes2023v,tong2022videomae,wang2023videomae} or contrastive pretraining~\cite{xu2021videoclip} paradigms to learn discriminative representations of video inputs. 
More recently, with the tremendous success of diffusion models, there is growing interest in learning generative models that exploit large-scale video priors. Such models achieve strong video synthesis results, exemplified by Sora~\cite{videoworldsimulators2024} and numerous follow-ups~\cite{wan2025wan,kong2024hunyuanvideo,opensora2,hong2022cogvideo,yang2024cogvideox}. 
While these models demonstrate very strong pixel synthesis performance, their internal representations are not well understood. In our work, we present a comprehensive study to understand how much 3D understanding these models possess. 

\paragraph{3D from Video} is a fundamental task in computer vision, traditionally tackled via Structure from Motion (SfM)~\cite{hartley2003multiple,schoenberger2016sfm,ozyecsil2017survey} and Multi-view Stereo (MVS)~\cite{furukawa2015multi} techniques. These classical methods rely on feature matching and cannot handle challenging cases (e.g. textureless regions, repetitive patterns, or wide baselines) well. Recent work instead turns to data-driven methods and propose strong feed-forward models for direct 3D prediction. This wave of research begins with pairwise models~\cite{wang2024dust3r,leroy2024grounding} and further extends to multi-view settings that improve accuracy and efficiency for large scenes~\cite{wang2025vggt,yang2025fast3r,tang2025mv}. These methods achieve better performance and robustness than classical approaches, yet it remains challenging for them to further scale up and to generalize to dynamic or real-world cluttered scenes given their reliance on annotated data.

To address this limitation, recent work considers using video priors from large video generative models. Existing work directly finetunes video generation models on 3D data, to achieve 3D control~\cite{he2024cameractrl,he2025cameractrl,bahmani2025ac3d,xu2024camco} or to simultaneously output 3D attributes~\cite{team2025aether,lu2025matrix3d,zhang2025world,huang2025jog3r,mai2025can,jiang2025geo4d}. Despite progress, small-scale finetuned video models still exhibit major artifacts of 3D inconsistency, especially on the data distinct from the fine-tuning data. To mitigate this, prior and concurrent works consider the usage of 1) explicit 3D memory~\cite{yu2024viewcrafter,gu2025diffusion,ren2025gen3c,huang2025voyager,wu2025video}; 2) post 3D optimization~\cite{sun2024dimensionx,liu2024reconx,chen2024v3d}; or 3) feedforward models on generations~\cite{liang2024wonderland,zhang2025spatialcrafter} to enhance 3D consistency. These efforts demonstrate the utility of video priors in sparse/single-view regimes by using video generators as frame extrapolators, yet the extent of 3D information already encoded in base video models remains unclear in quantitative terms.
Answering this question requires a model-agnostic framework that evaluates various models with quantitative metrics, which is what we pursue in this work.

\paragraph{Quantifying 3D awareness of visual foundation models} is an important research direction which helps the understanding of learned features and guides the development of scalable 3D world models. Early work in this area studies large image models and demonstrates the emergence of semantic correspondence in the feature space~\cite{amir2021deep,wei2022emergent,zhang2023tale,hedlin2023unsupervised}. To directly quantify 3D understanding, more recent works use 3D semantic (e.g. 3D-VQA, multi-view object recognition, semantic segmentation) or coarse estimation (e.g. relative depth) tasks to test the 3D understanding of visual foundation models~\cite{zhan2024general,zuo2025towards,man2024lexicon3d,fu2024blink,bonnen2024evaluating,sarkar2024shadows, li2025cue3d,ray2025satdynamicspatialaptitude}. 
Other work, such as VBench~\cite{huang2023vbench, huang2024vbench++, zheng2025vbench2} and WorldScore~\cite{duan2025worldscore}, focuses on benchmarking video generators and evaluates the 3D consistency of generated videos using off-the-shelf priors.
Instead of using coarse-grained or model-specific evaluation, Probe3D~\cite{el2024probing} and Feat2GS~\cite{chen2025feat2gs} consider dense probes to evaluate the 3D awareness of deep models and are more relevant to our work. However, their probes target image models, and their evaluation mainly focuses on depth/normal or multi-view consistency. In our study, we present a comprehensive study on video models by directly probing them with 3D attributes. We additionally show that indirect probes such as depth and multi-view consistency are not necessarily the best metrics to evaluate 3D awareness across different families of models.
\section{Approach}
We probe the 3D awareness of various VidFMs in this work. We define 3D awareness as the extent to which the underlying 3D structure and ego-motion can be recovered using frozen features extracted from 2D video. Under a fixed probe capacity and training set, stronger 3D awareness implies that a shallow readout attains a lower reconstruction error. Our method has two stages. First, we extract per-frame spatial feature maps by running each VidFM on video clips while keeping the VidFM parameters fixed. Second, we train a lightweight feedforward probe on these features to predict, for each sampled frame, (i) a dense 3D point map that represents the 3D coordinates of visible pixels in the coordinate system of the first frame, (ii) a dense depth map at a consistent scale with other frames, and (iii) the camera pose of each frame relative to the first frame; only the probe is optimized, not the VidFM. We primarily evaluate popular frontier video generation models~\cite{yang2024cogvideox,opensora2,wan2025wan,team2025aether}, and also include a self-supervised video encoder, V-JEPA~\cite{bardes2023v}, and two control models, DINOv2~\cite{oquab2023dinov2} and Fast3R~\cite{yang2025fast3r}, to contextualize our results.

\subsection{Feature Extraction} 
Given a video $\mathbf{V}\in\mathbb{R}^{T_v\times 3\times H_v\times W_v}$, we run each VidFM in frozen mode and extract, for every frame at $t$, a spatial feature map $\mathbf{F}_t\in\mathbb{R}^{C\times H_f\times W_f}$. For diffusion-based video generators, we extract features similar to DIFT~\cite{tang2023emergent}: we choose a denoising timestep $\tau$, add noise, perform a single denoising step, and read hidden activations from a specified network layer as features. We use an empty text embedding, and for image-to-video models we condition on the first frame. The layer index and $\tau$ are treated as hyperparameters and are fixed across experiments. For V-JEPA, DINOv2 and Fast3R, we run a standard forward pass and take the last-layer spatial features, which we empirically find to be the best-performing layer.

Different VidFMs often operate on different clip lengths. Several models we investigate are trained on fixed small context windows. To test them on longer videos, we split the input video $\mathbf{V}$ into short chunks for these models, by subsampling at fixed strides from beginning. We prepend the first frame to each chunk, so all chunks share the same first-frame reference. We also maintain a frame-to-feature index map $\pi(t)$ that records, for each raw frame at $t$, the corresponding chunk and local index. At probe time, based on the input frame indices $\{{t_i}\}_{i=1}^S$ and $\pi(t)$, we can gather the corresponding features $\{\mathbf{F}_{t_i}\}_{i=1}^S$ for all $S$ input frames. 

\subsection{3D Awareness Probe} 
\paragraph{Architecture.} We use a shallow transformer probe with alternating attention and three readout heads. For each input video, we take $S{=}4$ frames: the first video frame as the reference and three additional frames sampled with a minimum temporal gap of 5 frames. From the corresponding feature maps $\{\mathbf{F}_{t_i}\}_{i=1}^4$, we obtain per-frame tokens and apply four blocks of alternating-attention on top. Each alternating-attention block consists of a frame attention that mixes tokens within each frame and a global attention that mixes tokens across frames; this mirrors the VGGT design~\cite{wang2025vggt} but is much shallower. Three heads decode 3D outputs: two DPT heads produce dense point maps $\hat{\mathbf{X}}_{t_i}\in\mathbb{R}^{H_v\times W_v \times 3}$ (in the coordinate system of the first frame) and depth maps $\hat{\mathbf{D}}_{t_i}\in\mathbb{R}^{H_v\times W_v}$, similar to Probe3D~\cite{el2024probing} and VGGT. A camera head predicts the pose of each frame relative to the first frame. 

\paragraph{Loss.} We train the probe with a multi-task objective similar to VGGT:
$$
\mathcal{L}
= \lambda_{\text{pmap}}\mathcal{L}_{\text{pmap}}
+ \lambda_{\text{depth}}\mathcal{L}_{\text{depth}}
+ \lambda_{\text{cam}}\mathcal{L}_{\text{cam}},
$$
with $\lambda_{\text{pmap}}=\lambda_{\text{depth}}=\lambda_{\text{cam}}=1$ unless otherwise stated. For $\mathcal{L}_{\text{pmap}}$ and $\mathcal{L}_{\text{depth}}$, we use confidence-weighted $\ell_2$ losses between predicted point/depth maps and groundtruth point/depth maps. Note that the groundtruth scenes are normalized before loss calculation to remove scale ambiguity. For camera poses, we use a Huber loss between the predicted poses and groundtruth poses.

\section{Experiments}
\subsection{Evaluation}

\begin{table*}[t]
\centering
\small
\setlength{\tabcolsep}{3pt}      
\renewcommand{\arraystretch}{1.15} 
\begin{tabular}{l@{\hspace{10pt}}cccc@{\hspace{10pt}}cccc}
\toprule
& \multicolumn{4}{c}{CO3Dv2} & \multicolumn{4}{c}{DL3DV} \\
\cmidrule(lr){2-5} \cmidrule(lr){6-9}
Probed Feature &
Point Err(↓) & Depth Err(↓) & AUC@5 (↑) & AUC@30 (↑) &
Point Err(↓) & Depth Err(↓) & AUC@5 (↑) & AUC@30 (↑) \\
\midrule
\rowcolor{perframe}
DINOv2~\cite{oquab2023dinov2}      & 0.559 & 0.209 & 0.051 & 0.508 & 2.814 & 0.534 & 0.013 & 0.245 \\
\rowcolor{selfsup}
V-JEPA~\cite{bardes2023v}          & 0.439 & 0.214 & 0.076 & 0.619 & 1.576 & 0.613 & 0.076 & 0.558 \\
\rowcolor{videogen}
CogVideoX~\cite{yang2024cogvideox} & 0.485 & 0.231 & 0.051 & 0.569 & 1.748 & 0.608 & 0.061 & 0.486 \\
\rowcolor{videogen}
Aether~\cite{team2025aether}       & 0.501 & 0.249 & 0.054 & 0.571 & 1.566 & 0.574 & 0.067 & 0.527 \\
\rowcolor{videogen}
Open-Sora2.0~\cite{opensora2}      & 0.391 & 0.196 & 0.096 & 0.643 &
\secondbest{1.306} & \secondbest{0.445} & 0.115 & 0.607 \\
\rowcolor{videogen}
WAN2.1-14B~\cite{wan2025wan}       &
\secondbest{0.284} & \secondbest{0.151} & \secondbest{0.200} & \secondbest{0.736} &
\best{1.051} & \best{0.323} & \best{0.136} & \best{0.660} \\
\rowcolor{expert3d}
Fast3R~\cite{yang2025fast3r}       &
\best{0.262} & \best{0.145} & \best{0.272} & \best{0.769} &
1.379 & 0.514 & \secondbest{0.134} & \secondbest{0.637} \\
\bottomrule
\end{tabular}
\caption{\textbf{3D awareness benchmark results on CO3Dv2 and DL3DV.}
We evaluate \textbf{\videogentxt{video generators}},
\textbf{\selfsuptxt{self-supervised video encoders}},
\textbf{\experttxt{3D experts}},
and \textbf{\perframetxt{per-frame image models}}.
State-of-the-art video generators such as WAN2.1-14B and Open-Sora2.0
exhibit strong 3D awareness and outperform Fast3R on scenes. Point map errors have been multiplied by 10 for clarity.}
\label{tab:main-result}
\end{table*}

\paragraph{Datasets.} We perform experiments on CO3Dv2~\cite{reizenstein2021co3d} and DL3DV~\cite{ling2024dl3dv}. CO3Dv2 is an object-centric dataset consisting of turntable-type videos. We curate CO3Dv2 by filtering out sequences with heavy truncation or portrait-oriented videos that prevent forming border-less horizontal object-centric crops. The filtered split contains 11k videos in total. From each video, we sample consecutive frames as inputs to the feature extraction pipeline and use features from the first 76 frames during training. We adopt a 9:1 train\:test split at the video level and additionally create an ablation subset of 10 diverse categories (2.7\text{k} videos total) for ablation study. On the other hand, DL3DV contains large, cluttered scenes and is generally more challenging than CO3D. We use the first 6k splits and a 9:1 train\:test split by video. For both datasets, we run VGGT~\cite{wang2025vggt} to generate ground truth for every frame: dense point maps, depth, and camera poses. For point and depth maps, we also save the confidence maps, which are used in our losses. Unlike probe time where we only sample 4 frames from the video clips, we use all frames to generate the groundtruth. This leads to much more accurate annotations than the groundtruth originally provided by the datasets.

\paragraph{Metrics.} 
The main metrics to evaluate 3D awareness are errors of point map, pose, and depth predictions.
For point maps, we first normalize each scene to remove global scale, then align the predicted and ground-truth point clouds with the Umeyama algorithm~\cite{umeyama2002least}, and report the mean $\ell_2$ error.
For depth, we report the mean $\ell_2$ error after the same scene normalization.
For camera pose, we compute relative pose errors over all frame pairs: rotation error $e_R$ is the geodesic angle on $SO(3)$, and translation error $e_T$ is the angle between translation directions.
Accuracy at threshold $\theta$ is defined jointly as $\Pr[\max(e_R,e_T)\le\theta]$, i.e., both rotation and translation must be within $\theta$.
Following~\cite{wang2025vggt}, we report $\mathrm{AUC}@\Theta$, the area under this joint accuracy curve as $\theta$ sweeps from $0^\circ$ to $\Theta^\circ$ (e.g., $\Theta\in\{5,30\}$).

\begin{figure*}[t]
\centering
\includegraphics[width=\linewidth]{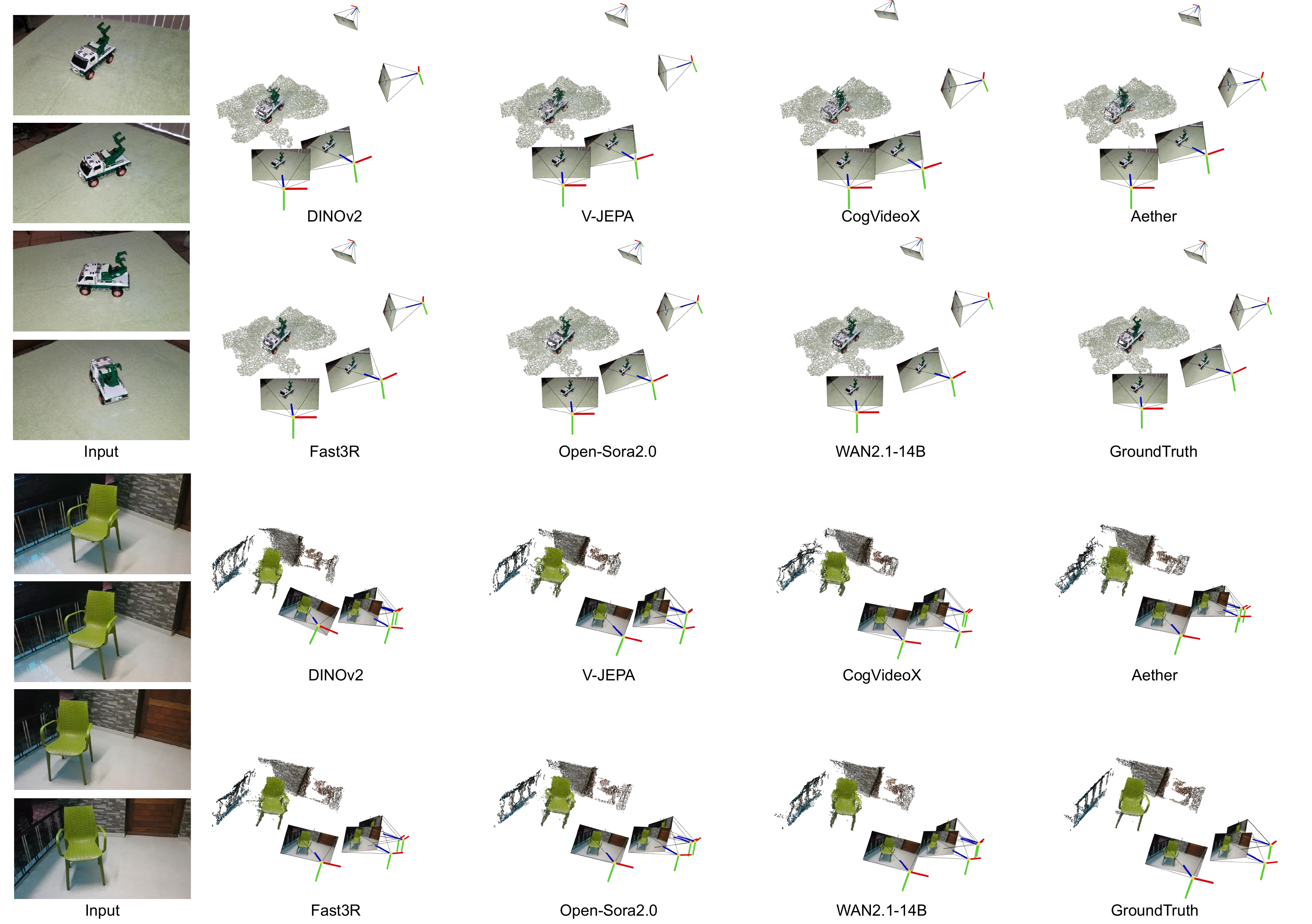}
\caption{\textbf{CO3Dv2 qualitative results.} For each scene, we show input frames and the unprojected 3D points prediction. Fast3R, WAN2.1-14B, and Open-Sora2.0 best preserve intricate details (e.g., the truck’s gripper) and reconstruct the overall structure.}
\label{fig:quali-co3d}
\end{figure*}

\begin{figure*}[t]
\centering
\includegraphics[width=\linewidth]{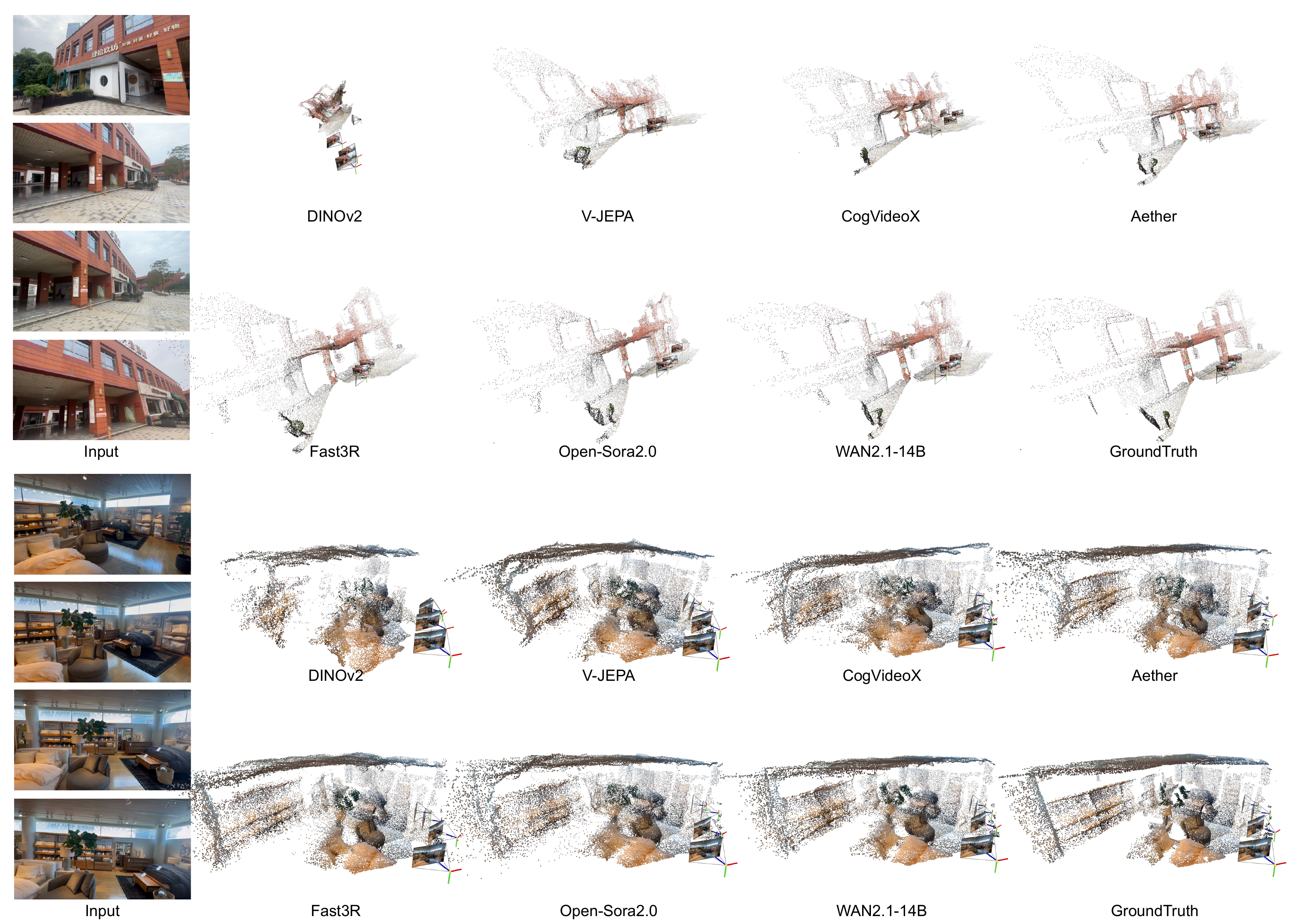}
\caption{\textbf{DL3DV qualitative results.} On this more challenging dataset, DINOv2 sometimes fails catastrophically. Top video generators often retain coherent geometry, where WAN2.1-14B produces the sharpest and most accurate point clouds overall.}
\label{fig:quali-dl3dv}
\end{figure*}

\paragraph{VidFM Baselines.} We evaluate various VidFM baselines, including video generators and self-supervised encoders. For generative models, WAN~\cite{wan2025wan} and Open-Sora2.0~\cite{opensora2} are the strongest open-weight generators we probe. We also probe CogVideoX~\cite{yang2024cogvideox}, an earlier work than WAN/Open-Sora2.0, and Aether~\cite{team2025aether}, which fine-tunes CogVideoX with 3D-aware objectives. All generative models here are latent diffusion models, which consist of a VAE that maps between raw videos and latents, and a denoiser that denoises the latents. For self-supervised models, we evaluate V-JEPA~\cite{bardes2023v}, a large self-supervised video encoder. 

\paragraph{Control groups.} A potential risk in our probe is that some 3D properties can already be estimated from raw videos. While relative rankings among VidFMs are still informative, if their probe performance is on par with direct 3D estimation from image features, the practical meaning of such rankings is compromised. To contextualize the results, we include two control baselines. \emph{Per-frame Image control:} we probe DINOv2 features extracted from each frame of the video. Since the features are extracted in isolation, any global 3D understanding of the video (e.g. 3D points in a common coordinate frame) must be induced by the probe rather than supplied by the backbone itself. To make the task well-posed, we append a reference-frame indicator token that marks the first frame; all losses, schedules, and hyperparameters mirror the VidFM setting. \emph{Native 3D control:} we probe features from Fast3R~\cite{yang2025fast3r}, a state-of-the-art model trained directly to predict 3D point maps from multi-view images. Because this model is optimized for the same target as our probe, probing it under the same probe architecture and supervision provides a strong reference. Meanwhile, CO3D is part of Fast3R's training sets but not DL3DV; this allows us to study the generalization behavior of its features. Together, the per-frame control (lower reference) and native-3D control (upper reference) contextualize VidFM results and ground our conclusions.

\begin{figure*}[t]
\centering
\begin{subfigure}[t]{0.32\linewidth}
\centering
\includegraphics[width=\linewidth]{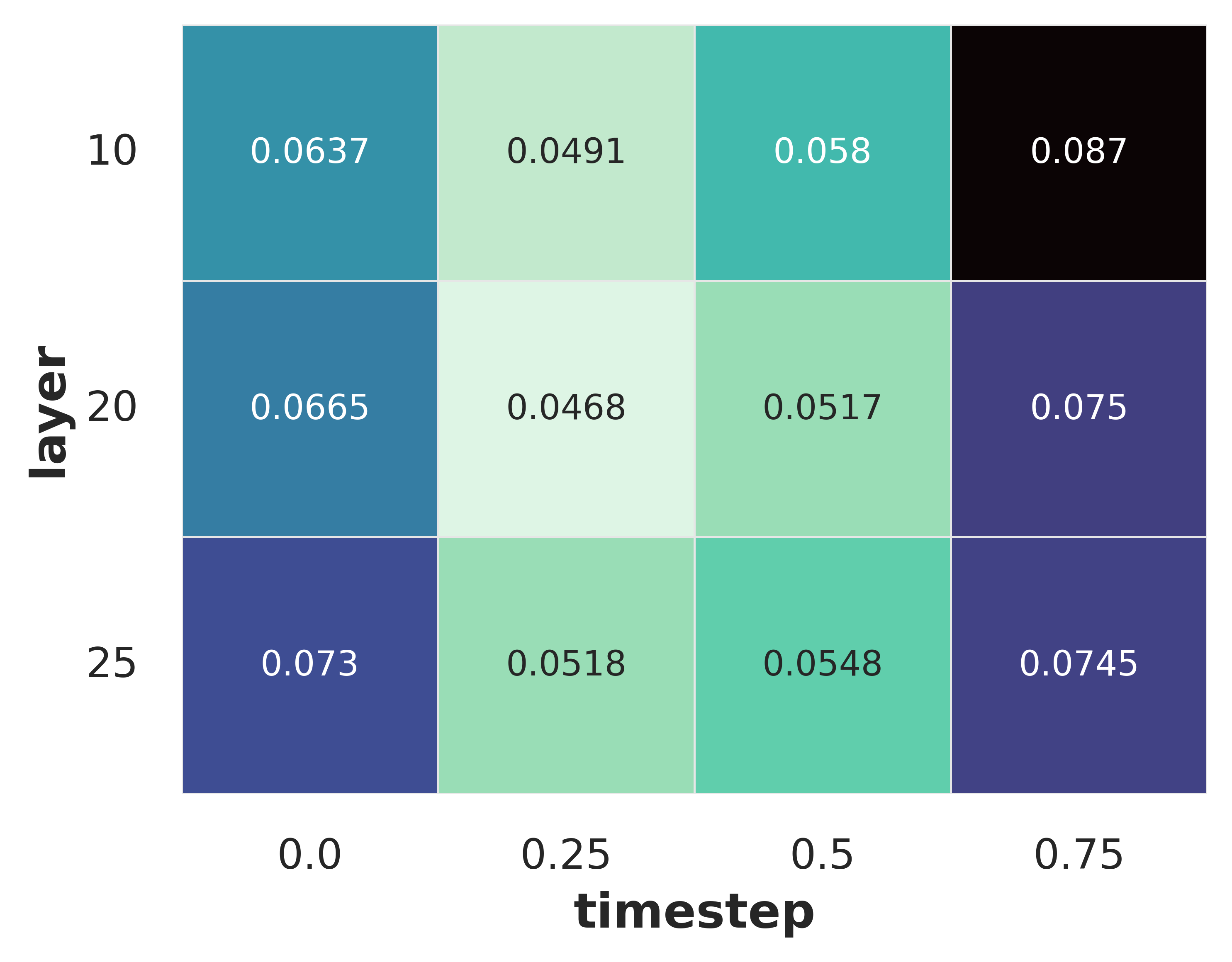}
\caption{WAN2.1}\label{fig:heat-wan}
\end{subfigure}\hfill
\begin{subfigure}[t]{0.32\linewidth}
\centering
\includegraphics[width=\linewidth]{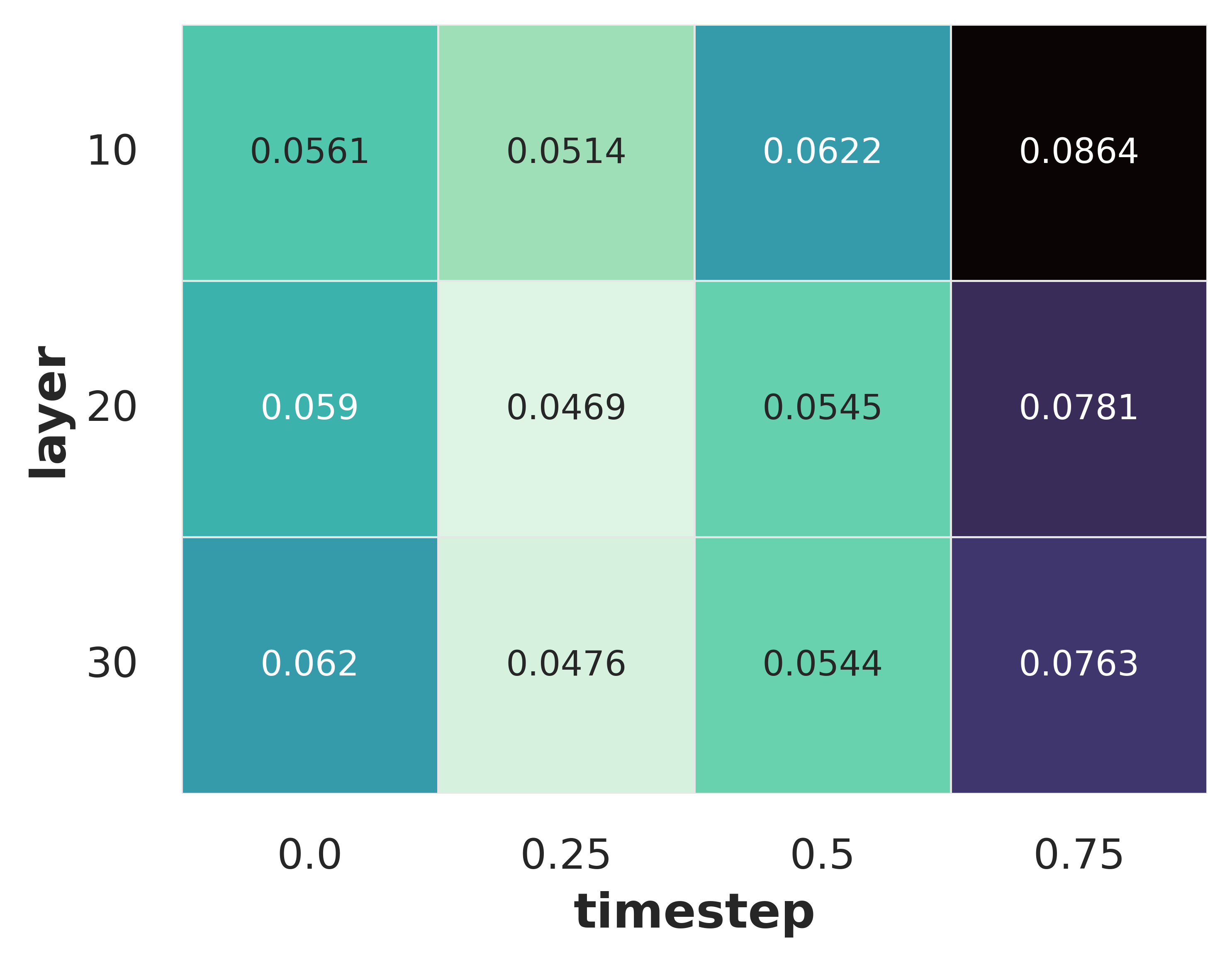}
\caption{Open\mbox{-}Sora2.0}\label{fig:heat-opensora}
\end{subfigure}\hfill
\begin{subfigure}[t]{0.32\linewidth}
\centering
\includegraphics[width=\linewidth]{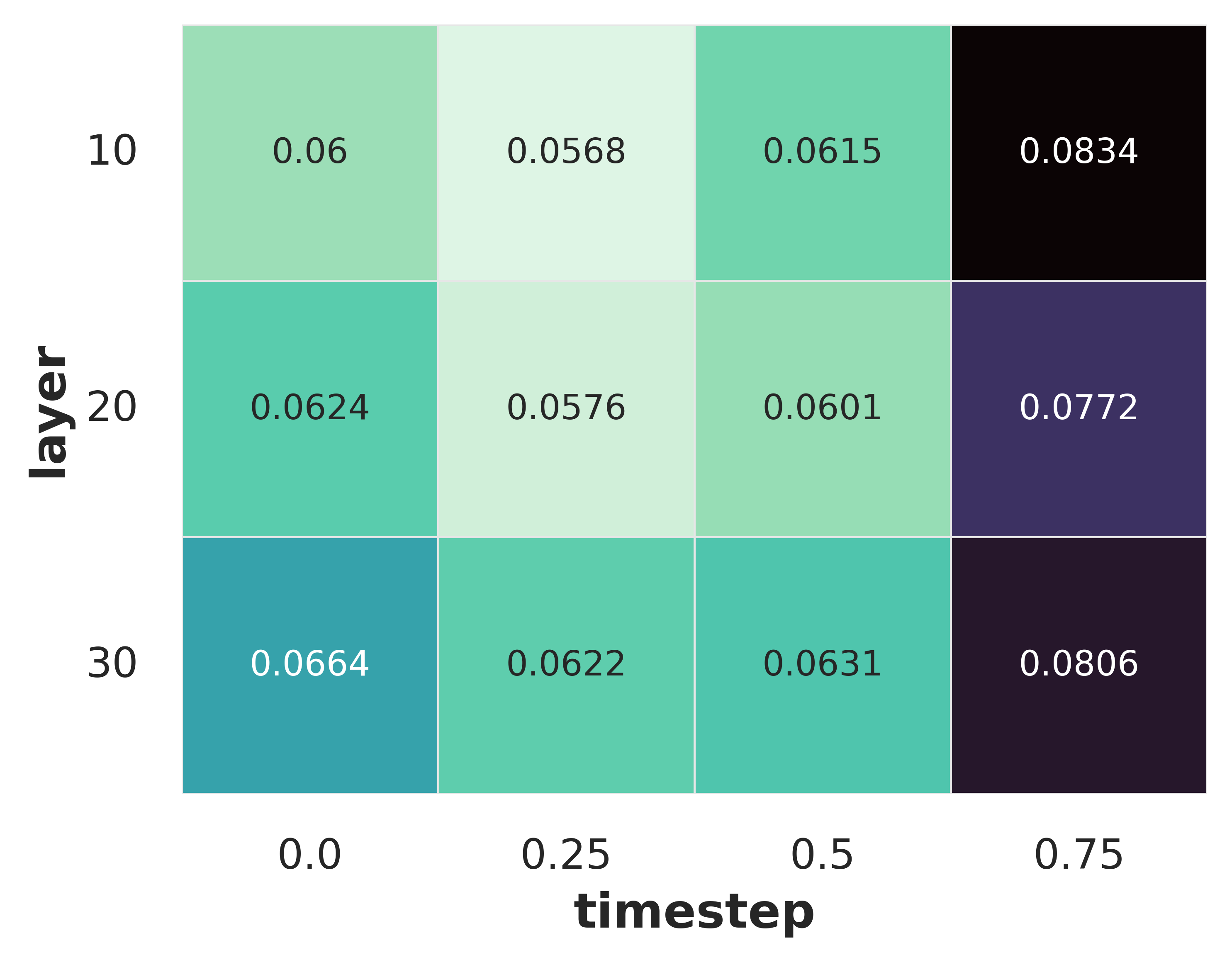}
\caption{CogVideoX}\label{fig:heat-cogvideo}
\end{subfigure}
\caption{\textbf{Layer--timestep ablations.} We show point-map error (lower is better) on the ablation data when probing different diffusion layers and denoising time steps. Best results are consistently from mid layers and early-but-not-first time steps.}
\label{fig:heat-layer-time}
\end{figure*}

\subsection{3D Awareness Benchmark}
We now present and analyze our results along the four axes defined earlier. We additionally analyze the relationship between our direct 3D probe and the multi-view evaluation from prior works in the supplementary material.

\paragraph{Extent: how does the 3D awareness of VidFMs compare to that of image models or specialized 3D models?}
\emph{Strong video diffusion models exhibit great 3D awareness even compared to 3D experts.} On CO3Dv2, WAN2.1-14B is second only to Fast3R across all metrics (e.g., Point 0.284 vs.\ 0.262, Depth 0.151 vs.\ 0.145, AUC@30 0.736 vs.\ 0.769, Table~\ref{tab:main-result} (left)). On DL3DV, which lies outside Fast3R’s training distribution, WAN2.1-14B surpasses Fast3R on all metrics (Point 1.051 vs.\ 1.379, Depth 0.323 vs.\ 0.514, AUC@30 0.660 vs.\ 0.637, Table~\ref{tab:main-result} (right)). Open-Sora2.0 is consistently strong as well, supporting the observation that state-of-the-art video generators yield features with universally strong 3D awareness across data domains.

\paragraph{Factor \#1: how does temporal reasoning impact 3D awareness?}
\emph{Effective temporal reasoning is critical to global 3D understanding.} Per-frame DINOv2 attains competitive depth on CO3Dv2 (0.209) but is significantly worse on global 3D understanding (Point 0.559, AUC@30 0.508) than all video models, including the self-supervised V-JEPA (Point 0.439, AUC@30 0.619). The key difference between image and video models is that the latter allows information exchange along the time axis. This gap in global 3D estimation widens on DL3DV (DINOv2 Point 2.814, AUC@30 0.245 vs. V-JEPA Point 1.576, AUC@30 0.558), whereas the depth estimation of DINOv2 remains competitive. The radar plots in Figure~\ref{fig:teaser} mirror this pattern: methods with explicit temporal reasoning produce polygons that expand along \emph{Point} and \emph{Pose}, not just \emph{Depth}.

\paragraph{Factor \#2: how does 3D finetuning impact 3D awareness?}
\emph{3D-aware fine-tuning does not always benefit.} Aether (fine-tuned from CogVideoX with 3D-aware objectives and conditions) indeed improves 3D awareness over CogVideoX on DL3DV (Point 1.566 vs.\ 1.748, Depth 0.574 vs.\ 0.608, AUC@30 0.527 vs.\ 0.486; Table~\ref{tab:main-result} (right)). However, on object-centric data, it is slightly worse than its base model (Point 0.501 vs.\ 0.485, Depth 0.249 vs.\ 0.231; Table~\ref{tab:main-result} (left)). Such discrepancy likely relates to the training data of Aether, which are mostly large synthetic scenes from games/simulators. This result suggests that 3D generative fine-tuning does have the potential to significantly improve 3D awareness, but how to avoid degraded generalization remains an interesting research direction.

\paragraph{Qualitative analysis.}
Figure~\ref{fig:quali-co3d} and Figure~\ref{fig:quali-dl3dv} align well with the ranking of 3D awareness in the quantitative tables. On CO3Dv2, Fast3R, WAN2.1-14B, and Open-Sora2.0 yield the most faithful and consistent reconstructions: thin structures and fine details (e.g., the gripper of the truck, the armrests and legs of the chair) remain sharp after unprojection, whereas other models exhibit noisy reconstructions and clear artifacts due to inconsistencies. On DL3DV, DINOv2 can fail catastrophically (e.g. the first building example, where the first view and the remaining views scarcely overlap), while top video generators often produce coherent point clouds. Overall, WAN2.1-14B delivers the sharpest and most accurate reconstructions, matching its lead in Table~\ref{tab:main-result} (right). Similarly, Aether demonstrates a clear improvement over CogVideoX qualitatively. Across both datasets, most failure cases concentrate around object boundaries.

\begin{table*}[h]
\centering
\resizebox{\textwidth}{!}{%

\begin{tabular}{lcccccccc}
\toprule
 & \multicolumn{4}{c}{CO3Dv2} & \multicolumn{4}{c}{DL3DV} \\
\cmidrule(lr){2-5}\cmidrule(lr){6-9}
Method & Point Err(↓) & Depth Err(↓) & AUC@5 (↑) & AUC@30 (↑)
       & Point Err(↓) & Depth Err(↓) & AUC@5 (↑) & AUC@30 (↑) \\
\midrule
Original VGGT~\cite{wang2025vggt}
  & 0.476 & 0.205 & 0.076 & 0.565
  & 2.751 & 0.518 & 0.058 & 0.363 \\
VidFM-VGGT
  & \textbf{0.289} & \textbf{0.145} & \textbf{0.178} & \textbf{0.718}
  & \textbf{1.034} & \textbf{0.319} & \textbf{0.183} & \textbf{0.686} \\
\bottomrule
\end{tabular}
}
\caption{\textbf{VidFM vs. DINO in VGGT.} Comparison between VGGT~\cite{wang2025vggt} (DINO features) and our VidFM-based variant using frozen WAN2.1-14B features on CO3Dv2 and DL3DV. Our model substantially improves all metrics, highlighting the advantage of video foundation model features for feedforward 3D reconstruction under limited 3D data.}
\label{tab:vggt_swap_feat}
\end{table*}

\subsection{Ablations}

\paragraph{Factor \#3: how does model size impact 3D awareness?}
On the ablation set, we further study whether models at larger scales produce more 3D-aware features. Given the limited availability of open-source checkpoints, we study the scaling of WAN and CogVideoX.
For WAN, scaling the model from 1.3B to 14B parameters significantly reduces point-map error from $0.0468$ to $0.0360$ on the ablation set (relatively ${-}23\%$). 
In contrast, CogVideoX slightly worsens in 3D awareness as parameters increase from 2B to 5B (from $0.0576$ to $0.0590$, relatively ${+}2\%$). This result suggests that parameter count alone does not guarantee stronger 3D awareness. We hypothesize that additional training data likely plays an important role here~\footnote{Unlike CogVideoX that mainly scales the architecture, WAN includes additional high-quality high-resolution data when scaling up.}.

\paragraph{Localization: in which network layers, and at which timesteps in diffusion models, is 3D information most concentrated?}
We ablate which diffusion \emph{layer} and \emph{timestep} yield the most 3D-aware features by sweeping over three network layers and four denoising timesteps. 
Across all the models we study, the optimum is consistent: \emph{mid-network layers} combined with an \emph{early-but-not-first} time step, are significantly better than other layers and time steps. For the choice of layers, the observation of \emph{mid-network layers} outperforming early or late layers is intuitive: in diffusion models, late layers are specialized to the per-frame RGB synthesis task, which suppresses high-level 3D-related features; whereas in too early layers, high-level features might not have formed yet. For the choice of time steps, in diffusion models, earlier time steps correspond to less noise added to the data or encoded feature. Considering the task of denoising, either too little or too much noise would lead to the degeneration of the task (i.e. either too easy or too hard) and make the features less useful. Comparing between early and late timesteps, early steps work better because the input signal is less corrupted by the noise. Overall, mid-layer and moderately early features strike a balance, retaining global 3D cues while being less influenced by the large noise added for denoising.

\subsection{VidFM Features for Feedforward 3D}
\label{subsec:feedforward}
Building on our previous analysis, we observe that features from video foundation models (VidFMs), especially video generative models such as WAN, are highly effective for 3D reconstruction. This raises a natural question: since current state-of-the-art feedforward 3D reconstructors like VGGT~\cite{wang2025vggt} rely on DINO features, how does the model perform with VidFM features such as WAN?

We investigate this question in our relatively small-data regime including DL3DV and CO3Dv2. We follow the same dataset split as in the previous experiments, and train (i) the original VGGT model from scratch, with DINO features optimized end-to-end, and (ii) an otherwise identical variant in which we replace DINO with frozen WAN2.1-14B features. Under a matched compute budget, we train both models to convergence and report the results in Table~\ref{tab:vggt_swap_feat}. On these benchmarks, our VidFM-based VGGT consistently outperforms the original VGGT by a large margin across all metrics. These results suggest that, when high-quality 3D supervision is limited to small datasets such as CO3Dv2 and DL3DV, it is preferable to use video model features rather than DINO features for feedforward 3D reconstruction.

\subsection{Limitations}
Our study relies on publicly released checkpoints rather than models trained under controlled conditions. 
Compute and data constraints prevent us from training video generators with precisely controlled variations at scale, so we cannot strictly attribute 3D-awareness differences to several factors of interest (e.g., data, training strategy).
In particular, to the best of our knowledge, there are no open-source models that provide multiple versions of checkpoints \emph{only} differing in the scale of training data; as a result, we cannot isolate the effect of data scale.
Meanwhile, due to resource constraints, we are unable to train large-scale 3D reconstruction models from scratch on massive datasets with VidFM features—an interesting direction for future work.

\section{Conclusion}
In this paper, we study the 3D awareness of video foundation models. Unlike prior work that focuses on image models and relies on 2.5D or optimization-based proxies, we probe video models using direct 3D prediction tasks. We find that state-of-the-art video generators exhibit strong, generalizable 3D awareness—even compared to domain experts. Our experiments demonstrate the importance of temporal reasoning for 3D understanding, and we examine how 3D fine-tuning, model scaling, and diffusion feature-extraction choices impact 3D awareness. Our experiments also show the promise of using VidFM features for 3D reconstruction in the limited-data regime. Beyond analysis, our work presents a 3D evaluation protocol and benchmark for existing video foundation models. We will publicly release our code, data, and weights, and we hope this work provides a solid step toward understanding and building scalable 3D world models.
{
    \small
    \bibliographystyle{ieeenat_fullname}
    \bibliography{main}
}

\clearpage
\appendix

\twocolumn[{
  \renewcommand\twocolumn[1][]{#1}
  \maketitlesupplementary
  \begin{center}
    \captionsetup{type=table}
    \small
    \setlength{\tabcolsep}{3pt}
    \renewcommand{\arraystretch}{1.15}
    \begin{tabular}{l@{\hspace{10pt}}ccc@{\hspace{10pt}}ccc}
      \toprule
      & \multicolumn{3}{c}{Original Probe} & \multicolumn{3}{c}{Smaller Probe} \\
      \cmidrule(lr){2-4} \cmidrule(lr){5-7}
      Probed Feature &
      Point Err(↓) & Depth Err(↓) & AUC@30 (↑) &
      Point Err(↓) & Depth Err(↓) & AUC@30 (↑) \\
      \midrule
      \rowcolor{perframe}
      DINOv2~\cite{oquab2023dinov2}      &  2.814 & 0.534 & 0.245 & 3.344 & 0.623 & 0.163 \\
      \rowcolor{selfsup}
      V-JEPA~\cite{bardes2023v}          &  1.576 & 0.613 &  0.558 &  1.707 & 0.657 & 0.505\\
      \rowcolor{videogen}
      WAN2.1-14B~\cite{wan2025wan}       &
      \best{1.051} & \best{0.323}  & \best{0.660} & \best{1.317} & \best{0.374} & \best{0.567}\\
      \rowcolor{expert3d}
      Fast3R~\cite{yang2025fast3r}        &
      1.379 & 0.514 & 0.637 & 1.551 & 0.572 & 0.549 \\
      \bottomrule
    \end{tabular}
    \captionof{table}{
    \textbf{Ablation on probe sizes.} We compare the 3D awareness evaluation results using our original probe against a smaller probe on DL3DV. The relative rankings and our conclusions remains unchanged despite the change of probe sizes.
    }
    \label{tab:probe_size_abl}
  \end{center}
  
}]

\begin{figure*}[t]
\centering

\textbf{(i) CO3Dv2 Results}\\[0.3em]
\begin{subfigure}[t]{0.32\linewidth}
\centering
\includegraphics[width=\linewidth,trim={0cm 0cm 0cm 0cm}, clip]{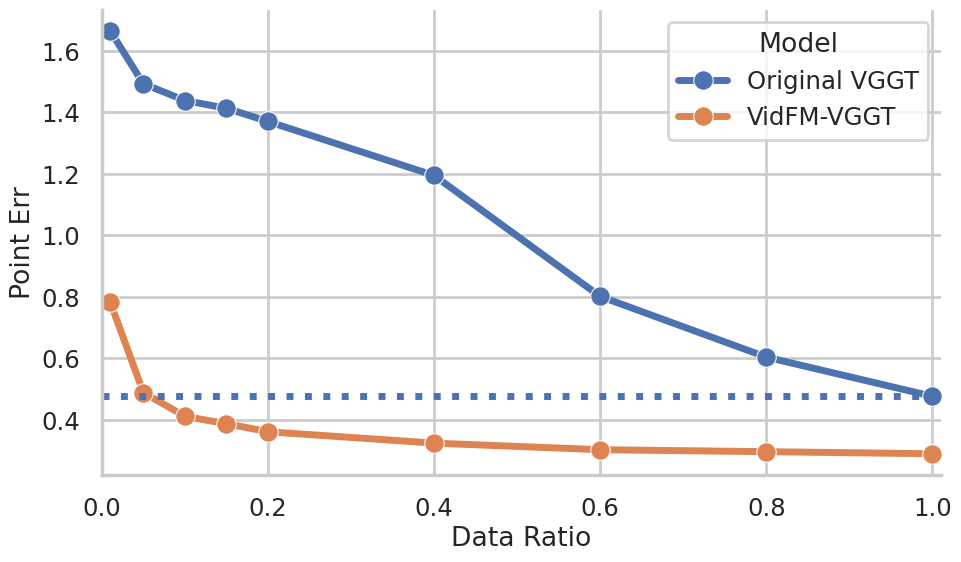}
\label{fig:data_scaling_CO3D_point}
\end{subfigure}\hfill
\begin{subfigure}[t]{0.32\linewidth}
\centering
\includegraphics[width=\linewidth,trim={0cm 0cm 0cm 0cm}, clip]{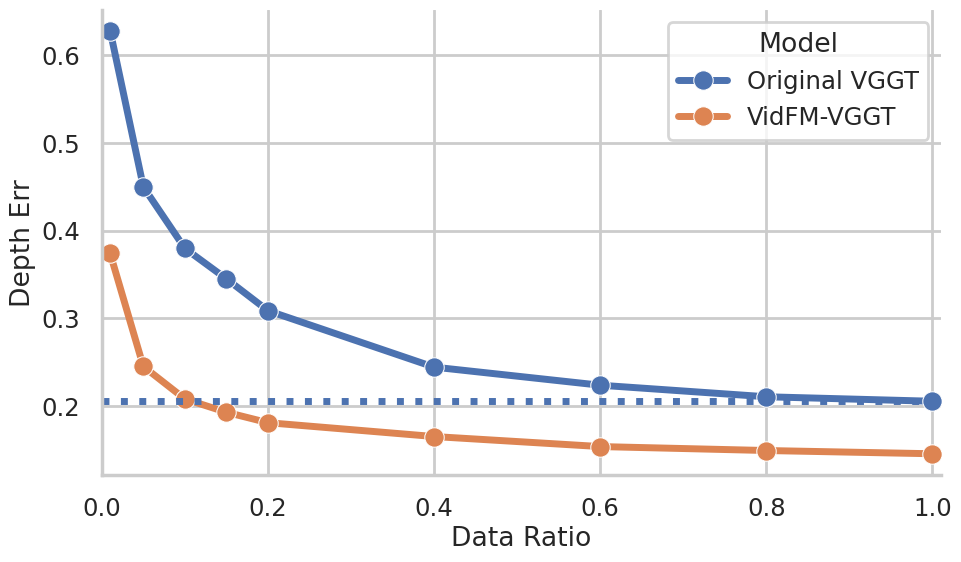}
\label{fig:data_scaling_CO3D_depth}
\end{subfigure}\hfill
\begin{subfigure}[t]{0.32\linewidth}
\centering
\includegraphics[width=\linewidth,trim={0cm 0cm 0cm 0cm}, clip]{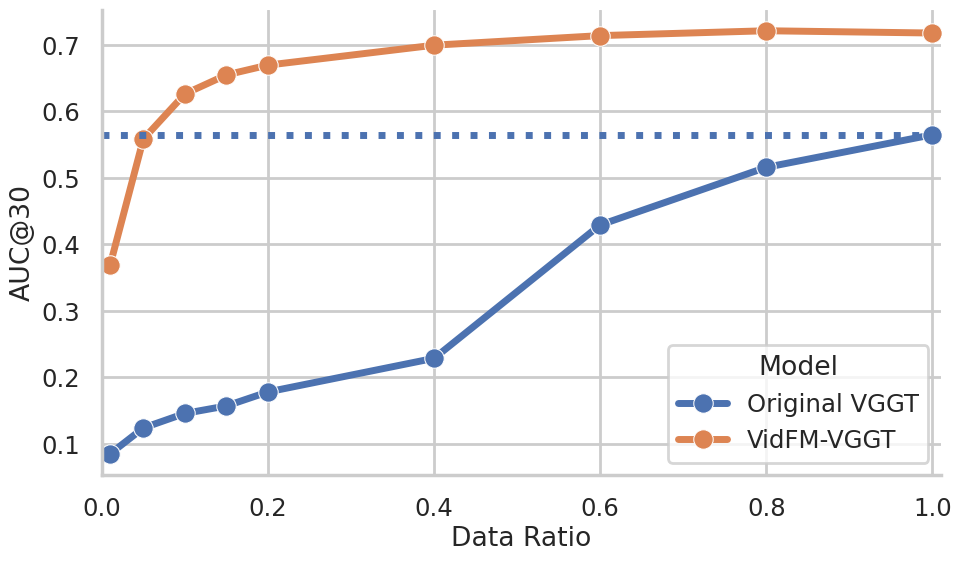}
\label{fig:data_scaling_CO3D_auc30}
\end{subfigure}

\vspace{1.2em}

\textbf{(ii) DL3DV Results}\\[0.3em]
\begin{subfigure}[t]{0.32\linewidth}
\centering
\includegraphics[width=\linewidth,trim={0cm 0cm 0cm 0cm}, clip]{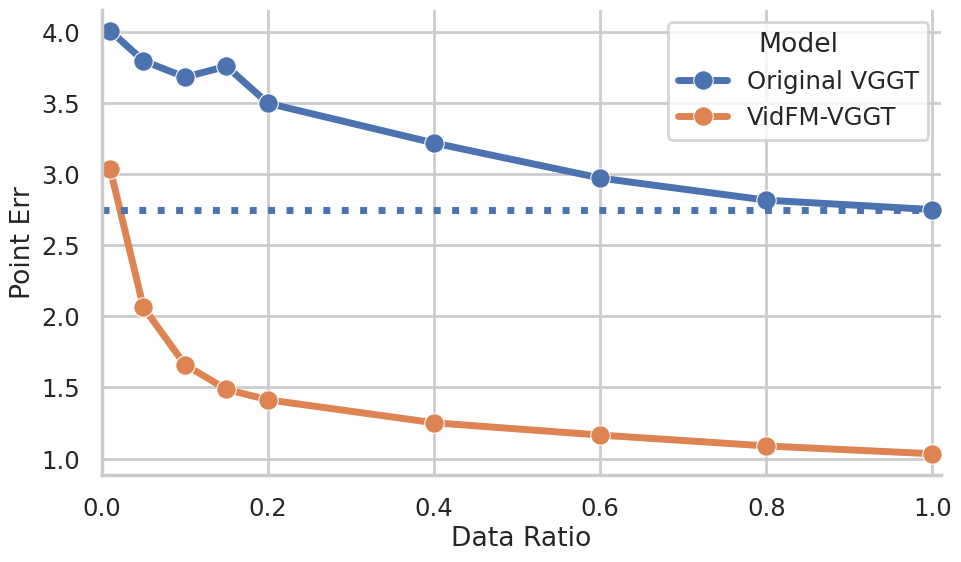}
\label{fig:data_scaling_dl3dv_point}
\end{subfigure}\hfill
\begin{subfigure}[t]{0.32\linewidth}
\centering
\includegraphics[width=\linewidth,trim={0cm 0cm 0cm 0cm}, clip]{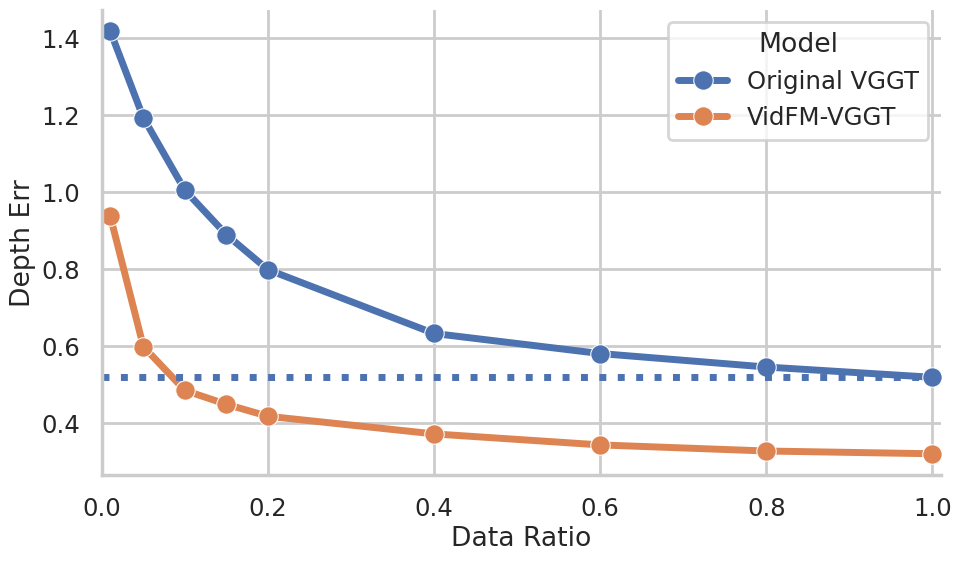}
\label{fig:data_scaling_dl3dv_depth}
\end{subfigure}\hfill
\begin{subfigure}[t]{0.32\linewidth}
\centering
\includegraphics[width=\linewidth,trim={0cm 0cm 0cm 0cm}, clip]{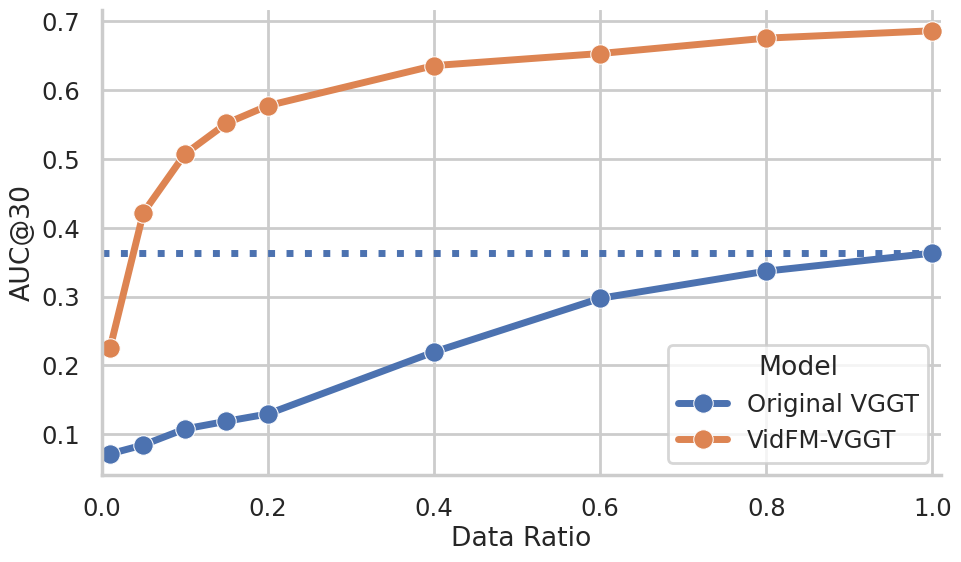}
\label{fig:data_scaling_dl3dv_auc30}
\end{subfigure}

\caption{\textbf{Data scaling on CO3Dv2 and DL3DV.} For each dataset we report point map error, depth error, and AUC@30 against the fraction of data used to train the model. The horizontal dashed line denotes the performance of the original VGGT trained with 100\% of the 3D data. VidFM VGGT typically outperforms this full-data baseline with less than 10\% of the 3D training data.}
\label{fig:data_scaling}
\end{figure*}

This supplementary material presents additional experiments and analyses on the 3D awareness of VidFMs.
In~\cref{sec:probe_size}, we study how probe size affects measured 3D awareness and show that our main conclusions are robust across probe capacities.
In~\cref{sec:data_size}, we extend our study in~\cref{subsec:feedforward} by showing how performance scales with the amount of 3D training data. We demonstrate that strong video generator features are especially beneficial for feedforward 3D reconstruction with limited 3D data or in challenging learning scenarios.
Finally, in Sec.~\ref{sec:mv_consistency}, we analyze the relationship between 3D probe performance and multi-view feature consistency. We find that cross-view correspondence alone can be a biased proxy for true 3D awareness, especially when comparing different model families.

\section{Ablation on Probe Size}
\label{sec:probe_size}

In our main experiments, we employ shallow probes with 4 layers and 1024 channels.
Here we evaluate whether our conclusions remain valid under even smaller probes.
We follow the same experimental protocol as the main paper, but use a significantly smaller probe by halving the model width from 1024 to 512. Table~\ref{tab:probe_size_abl} presents 3D awareness results for different-sized probes on DL3DV. We observe the relative performance remain stable across probe sizes; using a smaller probe does not affect our conclusions: features from state-of-the-art video generation models, e.g., WAN2.1-14B, exhibit strong 3D awareness compared to other model categories.

\section{Data Scaling for VidFM VGGT}
\label{sec:data_size}

In Table~\ref{tab:vggt_swap_feat} of the main paper, we compare the original VGGT with our variant that uses VidFM features. We show that using VidFM features significantly benefits feedforward 3D models under limited resources: under the same training data, VidFM-VGGT outperforms the original VGGT by a large margin.
We now extend this experiment by studying how performance changes with the amount of available training data. The scaling behaviors of both VGGT variants on CO3Dv2 and DL3DV are shown in Figure~\ref{fig:data_scaling}. In each plot, the dotted line denotes the performance of the original VGGT trained with 100\% of the 3D training data. Our VidFM-VGGT typically surpasses the full-data baseline with only less than 10\% of the training data across all metrics. Such contrast suggests that it is possible to induce strong 3D understanding from video features with a tiny fraction of 3D data, especially when compared to the commonly used image features. 
Thus, strong video generator features are particularly valuable in low-data settings. The gap is especially large on DL3DV, where the scenes are much more diverse and cluttered. This indicates that strong video generator features substantially benefit 3D learning in diverse and challenging data domains. Due to the availability of compute and data, we are not able to extrapolate our curves to the scale of original VGGT's training set, which pools most available 3D data. Such extrapolation will be an interesting future direction.

\section{Analysis on Multi-view Consistency}
\label{sec:mv_consistency}
We study how a model's 3D probe performance relates to multi-view consistency, which prior works often consider as a proxy for 3D awareness~\cite{el2024probing}. 

\begin{figure}[t]
\centering
\includegraphics[width=\linewidth]{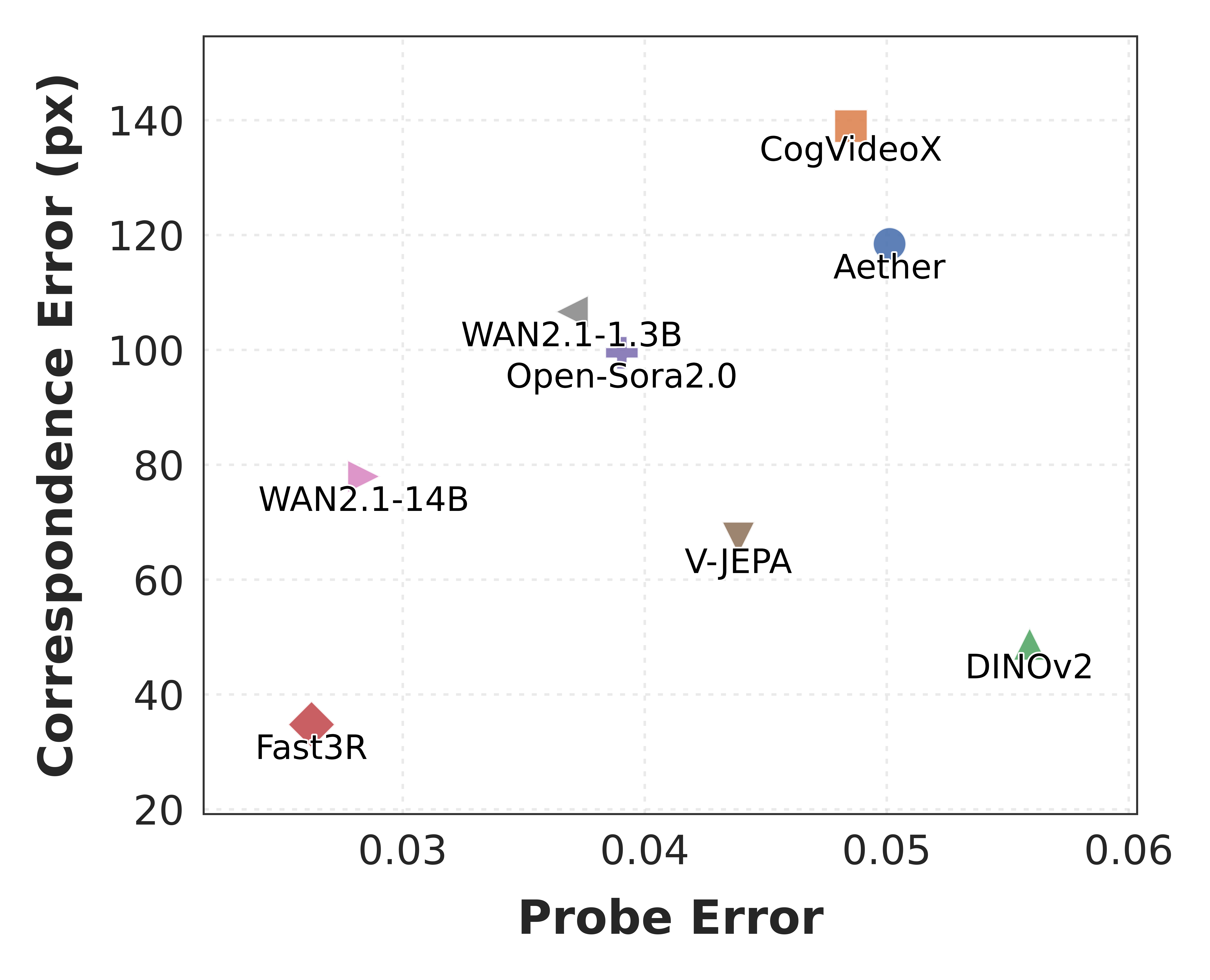}
\caption{\textbf{3D awareness vs.\ multi-view consistency.}
Scatter plot of \emph{3D Probe Error} (lower is better) versus \emph{Cross-view Correspondence Error} (lower is better). Within the family of video diffusion models, the 3D probe error positively correlates with the multi-view correspondence error. DINOv2 and V-JEPA achieve great multi-view correspondence, while performing significantly worse in 3D probing experiments. This suggests that cross-view feature similarity may not be a sufficient proxy for measuring 3D awareness, especially when comparing across families of models.}
\label{fig:corr-3dconsistency}
\vspace{-2ex}
\end{figure}

\paragraph{Measuring multi-view consistency.}
To quantify multi-view consistency, we measure the \emph{cross-view correspondence error} of different VidFM features. 
Cross-view correspondence error is defined as the pixel distance between the predicted correspondence and groundtruth correspondence.
To obtain groundtruth correspondence, we sample a random anchor view $A$ and a set of pixels within this view. We then reproject these pixels to another view $B$ using ground-truth 3D, and record their locations if they are not occluded. 
To obtain predicted correspondence, we use the standard nearest neighbor query in feature space: for each anchor points in view $A$, we retrieve the top-1 nearest neighbor in view $B$ based on the VidFM features. We then compute the average Euclidean pixel distance between the predicted correspondence and groundtruth correspondence. We use this mean distance as our measure of multi-view feature consistency, reported as cross-view correspondence error.

\paragraph{Correlation between 3D probe and multi-view consistency.}
Figure~\ref{fig:corr-3dconsistency} plots 3D probe error (x axis; lower is better) against cross-view correspondence error in pixels (y axis; lower is better). We perform this analysis on CO3Dv2, where the probe error is the point error reported in Table~\ref{tab:main-result} in the main paper.
Among video diffusion models, we observe a positive correlation, where lower probe error accompanies lower correspondence error. CogVideoX is the worst on both axes, Open-Sora2.0 and WAN2.1-1.3B are intermediate, and WAN2.1-14B is the best (bottom-left). 
By contrast, feedforward models (Fast3R, V-JEPA, DINOv2) lie \emph{below} the diffusion models. At a comparable probe error, they show better multi-view consistency. 
Within feedforward models, DINOv2 achieves particularly strong multi-view consistency, yet performs poorly at inferring global 3D properties from its features. We now discuss possible reasons for these observed discrepancies.

\paragraph{Comparison: diffusion models vs.\ feedforward models.}
Diffusion models exhibit worse multi-view feature consistency than feedforward models at the same level of 3D awareness. 
This follows from how diffusion features are extracted: noise is injected into the VAE features and a single denoising step is performed to estimate the noise or velocity. This not only makes features noisy at locations where large noise is added, but the underlying representation also includes features specifically tailored to denoising, which is affected by the random noise. 
Consequently, two pixels corresponding to the same 3D point across frames can carry different features, leading to feature discrepancies that suppress the raw feature consistency even when the underlying 3D structure is well-recoverable by shallow probes.

\paragraph{Comparison: video models vs.\ image models.}
DINOv2 attains especially strong multi-view consistency, surpassing even the self-supervised video encoder V-JEPA. 
We hypothesize that in video models some channels correlate with local motions at the current frame; pixels corresponding to the same 3D point may exhibit different local motions across frames. 
In this way, while video models encode richer temporal information that aids 3D decoding, their features can appear less “consistent” under nearest-neighbor matching. Such factor makes feature consistency alone a potentially biased evaluation for measuring 3D awareness.

\end{document}